\documentclass{svmult}

\usepackage{mathptmx}       
\usepackage{helvet}         
\usepackage{courier}        
\usepackage{type1cm}        

\usepackage{makeidx}         
\usepackage{graphicx}        
\usepackage{multicol}        
\usepackage[bottom]{footmisc}
\usepackage{amsmath,amssymb}

\makeindex             

\begin{document}
\title{Self-Motion of the 3-\underline{PP}PS Parallel Robot with Delta-Shaped Base}

\author{\textbf{D. Chablat$^1$, E. Ottaviano$^2$, S. Venkateswaran$^3$}}
\authorrunning{D. Chablat, E. Ottaviano, S. Venkateswaran}
\institute{
$^1$CNRS, Laboratoire des Sciences du Num\'erique de Nantes, UMR CNRS 6004, France, 
\email{Damien.Chablat@cnrs.fr},\\
$^2$DICeM, University of Cassino and Southern Lazio, via G. Di Biasio 43, 03043, Cassino (FR), Italy, 
\email{ottaviano@unicas.it}, \\
$^3$Ecole Centrale de Nantes, Laboratoire des Sciences du Num\'erique de Nantes, UMR CNRS 6004, France, 
\email{swaminath.venkateswaran@ls2n.fr}\\
}

\maketitle    
\abstract{This paper presents the kinematic analysis of the 3-\underline{PP}PS parallel robot with an equilateral mobile platform and an equilateral-shaped base. 
Like the other 3-\underline{PP}PS robots studied in the literature, it is proved that the parallel singularities depend only on the orientation of the end-effector.  
The quaternion parameters are used to represent the singularity surfaces. The study of the direct kinematic model shows that this robot admits a self-motion of the Cardanic type. This explains why the direct kinematic model admits an infinite number of solutions in the center of the workspace at the ``home'' position but has never been studied until now. 
}
\section{Introduction}
It has been shown that by applying simplifications in parallel robot design parameters, self-motions may appear \cite{Husty:1994,Karger:2002,Wohlhart:2002}. For example, Bonev et al. demonstrated that all singular orientations of the popular 3-RRR spherical parallel robot design (known as the Agile Eye) correspond to self motions \cite{Bonev:2006}, but arguably this design has the ``best'' spherical wrist. The Cardanic motion can be found as self-motion for two robots in literature, the 3-\underline{R}PR parallel robot \cite{Chablat:2006} and the PamInsa robot \cite{Briot:2008}.

Most of the examples for a fully parallel 6-DOF manipulator can be categorized by the type of their six identical serial chains namely UPS \cite{Merlet:2006, Pierrot:1998, Corbel:2008, sto:1993, ji:1999}, RUS \cite{hon:1997} and PUS \cite{hun:1983}.  However for all these robots, the orientation of the workspace is rather limited due to the interferences between the legs.

To solve this problem, new parallel robot designs with six degrees of freedom appeared recently having only three legs with two actuators per leg. The Monash Epicyclic-Parallel Manipulator (MEPaM), called 3-\underline{PP}PS  is a six DOF parallel manipulator with all actuators mounted on the base \cite{Chen:2012}. Several variants of this robot have been studied were the three legs are made with three orthogonal prismatic joints and one spherical joint in series. The first two prismatic joints of each legs are actuated. In the first design, the three legs are orthogonal \cite{Caro:2012}. For this design, the robot can have up to six solutions to the Direct Kinematic Problem (DKP) and is capable of making non-singular assembly mode change trajectories \cite{Caro:2012}. In \cite{Chen:2012}, the robot was made with an equilateral mobile platform and an equilateral-shape base. In \cite{Chablat:2017,Chablat:2018}, the robot was designed with an equilateral mobile platform and a U-shaped base. For this design, the Direct Kinematic Model (DKM) is simple and can be solved with only quadratic equations. 

The common point of these three variants is that the parallel singularity is independent of the position of the end-effector. This paper presents the self-motion of the 3-\underline{PP}PS parallel robot derived from \cite{Chen:2012} with an equilateral mobile platform and an equilateral-shaped base. 

The outline of this article is as follows. In the next section, the manipulator architecture as well as its associated constraint equations are explained. Followed by that, calculations for parallel singularities as well as self-motions are presented. The article then ends with conclusions.
\section{Mechanism Architecture}
The robot under study is a simplified version of the  MEPaM that has been developed at the Monash University \cite{Chen:2012,Caro:2010}. This architecture is derived from the 3-\underline{PP}SP that was introduced earlier \cite{Byun:1997}.  
\subsection{Geometric Parameters}
\begin{figure}[!ht]
\begin{center}
\includegraphics[scale=.45]{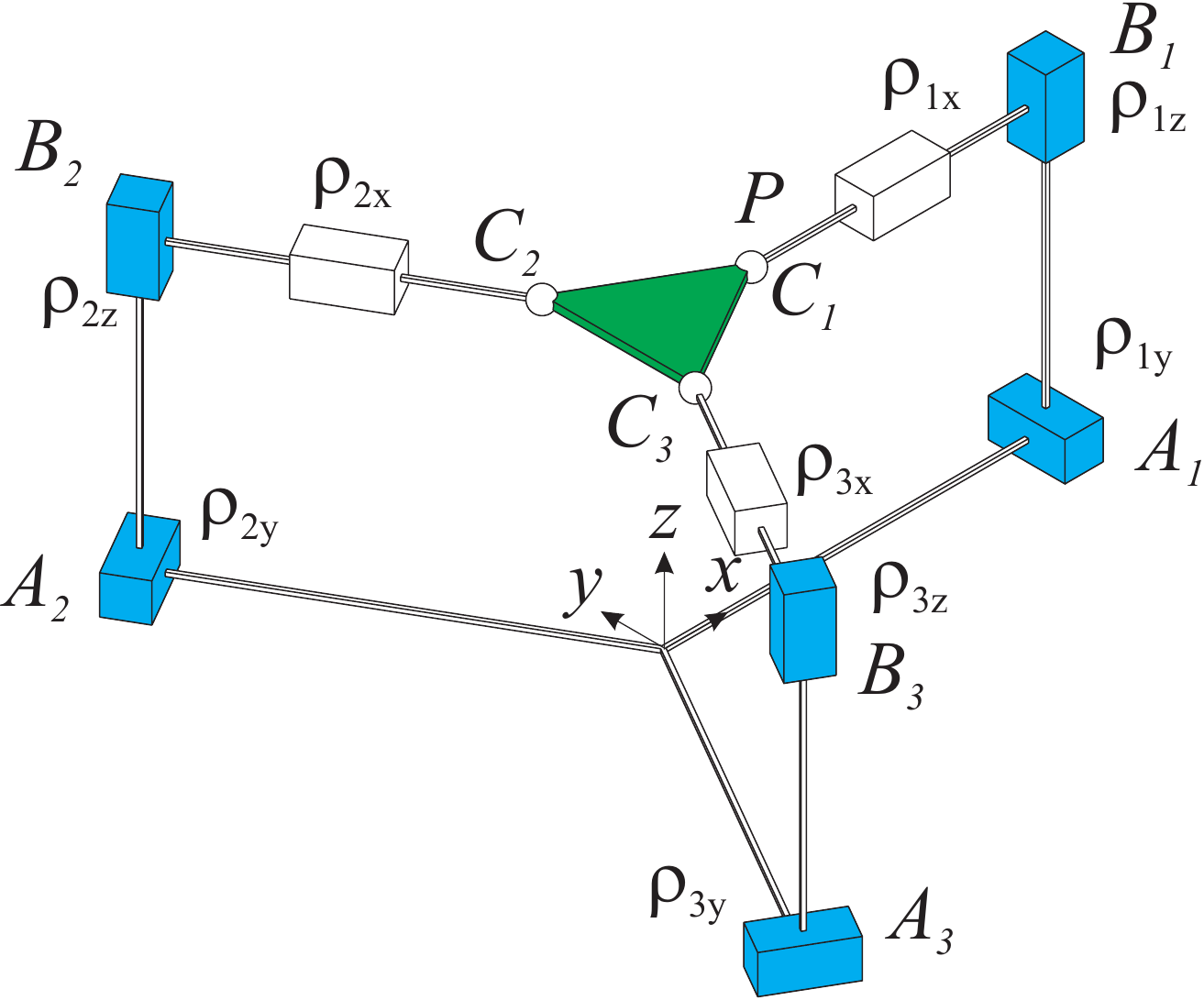}
\end{center}
\caption{A scheme for the 3-\underline{PP}PS parallel robot and its parameters in its ``home'' position with the actuated prismatic joints in blue, the passive joints in white and the mobile platform drawn in green with $x=1/\sqrt{3}, y=0, z=0, q_1=1, q_2=0, q_3=0, q_4=0$}
\label{Fig:Robot}
\end{figure}

For this parallel robot, the three legs are identical and it consists of two actuated prismatic joints, a passive prismatic joint and a spherical joint (Figure~\ref{Fig:Robot}). The axes of first three joints of each leg form an orthogonal reference frame. 

The coordinates of the point $C_1$ are $\rho_{1x}$, $\rho_{1y}$ and $\rho_{1z}$, wherein the last two are actuated. We assume an origin $A_i$ for each legs and the equations are given by:
\begin{gather}
 {\bf A}_1=[ 2, 0, 0]^T \\
 {\bf A}_2=[-1,~\sqrt {3},0]^T \\
 {\bf A}_3=[-1,-\sqrt {3},0]^T
\end{gather}
The coordinates of  $C_2$ and $C_3$ are obtained by a rotation around the $z$ axis by $2\pi/3$ and $-2\pi/3$ respectively and the equations are as follows: 
\begin{gather}
 {\bf C}_1=[ \rho_{1x}, \rho_{1y}, \rho_{1z}]^T \\
 {\bf C}_2=[-\rho_{2x}/2-\sqrt {3}\rho_{2y}/2, \sqrt {3}\rho_{2x}/2-\rho_{2y}/2,\rho_{2z}]^T \\
 {\bf C}_3=[-\rho_{3x}/2+\sqrt {3}\rho_{3y}/2,-\sqrt {3}\rho_{3x}/2-\rho_{3y}/2,\rho_{3z}]^T
\end{gather}
Three locations are now written to describe the mobile platform in the moving frame for an equilateral triangle whose edge lengths are set to one.
\begin{gather}
{\bf V}_1 = [0,0,0]^T \\
{\bf V}_2 = [-\sqrt{3}/2, 1/2,0]^T \\
{\bf V}_3 = [-\sqrt{3}/2,-1/2,0]^T
\end{gather}
Generally in the robotics community, Euler or Tilt-and-Torsion angles are used to represent the orientation of the mobile platform \cite{Caro:2012}. These methods have a physical meaning, but there are singularities to represent certain orientations. 
The unit quaternions give a redundant representation to define the orientation but at the same time it gives a unique definition for all orientations. The rotation matrix $\bf R$ is described by:
\begin{equation}
{\bf R}=
 \left[ \begin {array}{ccc} 
  2 q_1^2+2 q_2^2-1 & -2 q_1 q_4+2 q_2 q_3 & 2 q_1 q_3 + 2 q_2 q_4  \label{EQ:r}\\ 
  2 q_1 q_4+2 q_2 q_3 & 2 q_1^2 + 2 q_3^2-1 & -2 q_1 q_2+2 q_3 q_4 \\ 
 -2 q_1 q_3+2 q_2 q_4 & 2 q_1 q_2+2 q_3 q_4 & 2 q_1^2 + 2 q_4^2-1
\end {array} \right] 
\end{equation}
\noindent Here $q_1 \geq 0$ and $q_1^2+q_2^2+q_3^2+q_4^2=1$. 
We can write the coordinates of the mobile platform using the previous rotation matrix as:
\begin{equation}
  {\bf C}_i = {\bf R} {\bf V}_i + {\bf P} \quad {\rm where } \quad {\bf P}= [x, y, z]^T
\end{equation}
Thus, we can write the set of constraint equations with the position of $C_i$ in the both reference frames by the relations:
\begin{gather}
\rho_{1y}=y\\
\rho_{1z}=z\\ 
\left( 2 q_1 q_4-x \right) \sqrt {3}+2 {q_1}^{2}+3 {q_2}^{2}-{q_3}^{2}-y-2 \rho_{2y}=1\\
-\sqrt {3}q_1 q_3+\sqrt {3}q_2 q_4-q_1 q_2-q_3 q_4+\rho_{2z}=z\\
\left( 2 q_1 q_4+x \right) \sqrt {3}-2 {q_1}^{2}-3 {q_2}^{2}+{q_3}^{2}-y-2 \rho_{3y}=-1\\
-\sqrt {3}q_1 q_3+\sqrt {3}q_2 q_4+q_1 q_2+q_3 q_4+\rho_{3z}=z
\end{gather}
\subsection{Constraint equations}
The main problem is to find the location of the mobile platform by looking for the values of the passive prismatic joints $[ \rho_{1x}, \rho_{1y}, \rho_{1z}]$ as proposed in \cite{Parenti-Castelli:1990}. The distances between any couple of points $C_i$ are given by:
\begin{equation}
||{\bf C}_1-{\bf C}_2|| = ||{\bf C}_1-{\bf C}_3|| = ||{\bf C}_2-{\bf C}_3|| = 1
\end{equation}
This method is also used by \cite{Chen:2012} for the 3-\underline{PP}PS by using the dialytic elimination \cite{Angeles:2007}. A fourth degree polynomial equation with complicate coefficients is then obtained. However, no self-motion was detected. 

\begin{figure}[!ht]
\begin{center}
\includegraphics[scale=0.45]{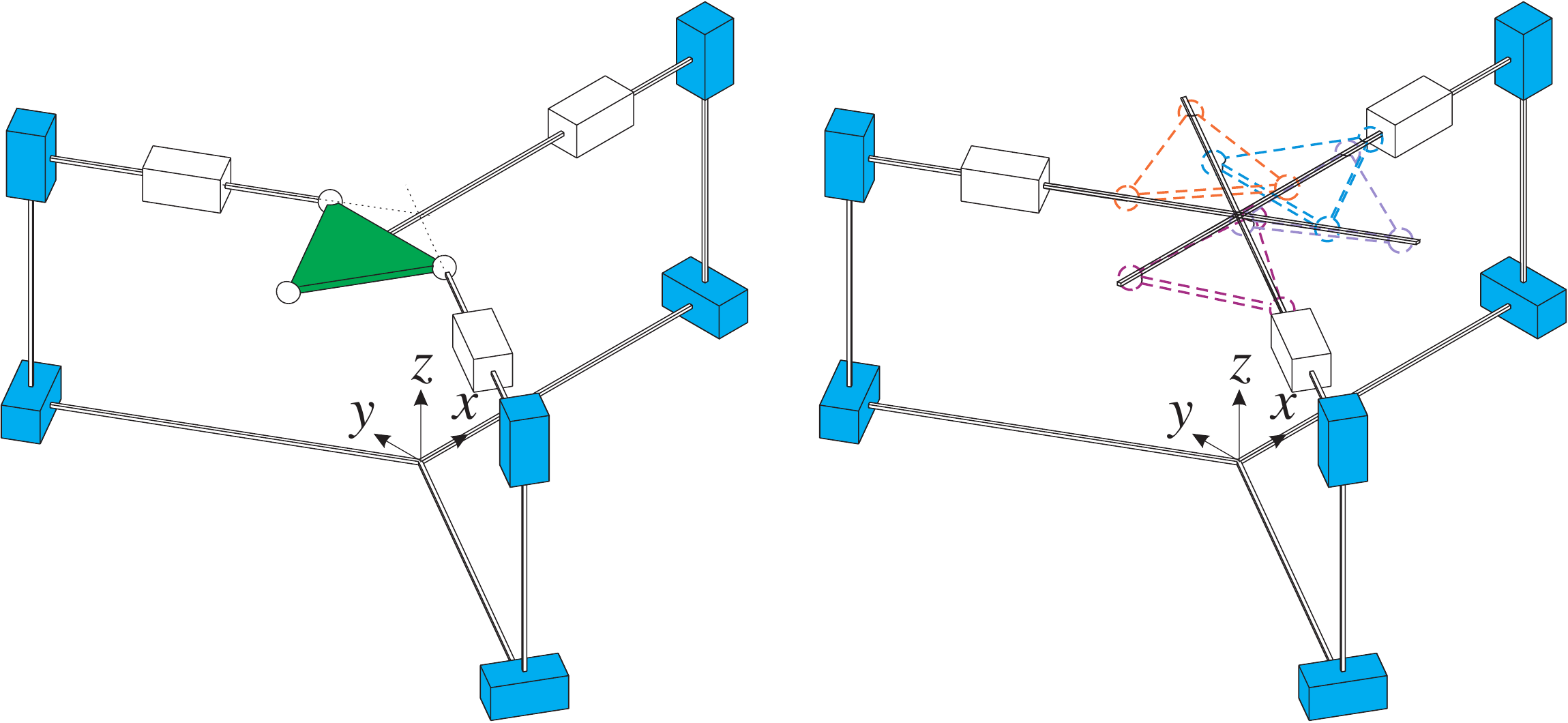}
\end{center}
\caption{Set of postures to describe the Cardanic motion starting from the ``home'' position}
\label{Fig:Singularity_self_motion}
\end{figure}

Unfortunately, when we want to solve the DKM for the ``home'' position, i.e. $\rho_{iy}=0$ and $\rho_{iz}=0$ and $i=1,2,3$, we have an infinite number of solutions, which correspond to the self-motion. This result remains the same if we set $\rho_{1z}=\rho_{2z}=\rho_{3z}$. Assuming that this motion is in a plane parallel to the plane (0xy), we write the system coefficients for $\rho_{iz}=0$ with $i=1,2,3$. The fourth degree polynomial then degenerates and a quadratic  polynomial equation is obtained which is of the form:
\begin{gather}
9 (\rho_{1y}+\rho_{2y}+\rho_{3y})^2 \rho_{1x}^2 + \nonumber \\
6 \sqrt{3}(\rho_{1y}+\rho_{2y}+\rho_{3y})^2 (\rho_{2y}-\rho_{3y}) \rho_{1x} + \\
(\rho_{1y}+\rho_{2y}+\rho_{3y})^2(\rho_{1y}^2+2 \rho_{1y} \rho_{2y}+2 \rho_{1y} \rho_{3y}+4 \rho_{2y}^2-4 \rho_{2y} \rho_{3y}+4 \rho_{3y}^2-3)=0 \nonumber
\end{gather}
Here all the terms of the equation cancel out each other when $(\rho_{1y}+\rho_{2y}+\rho_{3y} = 0)$. With these conditions, the robot becomes similar to the 3-\underline{R}PR parallel robot for which its self-motion produces a Cardanic motion. Since the self-motion  corresponds to singular configurations, the objective now will be to locate these singularities with respect to other singularities. Geometrically, the self-motion has been described when the motion axes of passive joints intersect at a single point and  an angle of $2\pi/3$ is formed between each axis.
An example is shown in Fig.~\ref{Fig:Singularity_self_motion} where the green mobile platform is one assembly mode obtained in the ``home'' position. This phenomenon has already been characterized for the 3-\underline{R}PR parallel robot \cite{Chablat:2006}  and the PamInsa robot  in \cite{Briot:2008}.
To validate this result, we use the Siropa library programmed under Maple and its ``InfiniteEquations'' function \cite{Siropa}. Three condition are obtained which are $(\rho_{1y}+\rho_{2y}+\rho_{3y} = 0)$, $\rho_{1z}=\rho_{2z}$ and $\rho_{2z}=\rho_{3z}$.
\section{Singularity Analysis}
The singular configurations of the 3-PPPS robot have been studied in several articles with either a parametrization of orientations using Euler angles or Quaternions \cite{Chablat:2017}. Serial and parallel Jacobian matrices can be computed by differentiating the constraint equations with respect to time  \cite{Gosselin:1990,Sefrioui:92,Chablat:1998}. These Jacobian serial and parallel matrices must satisfy the following relationship
\begin{equation}
  {\bf A t + B \dot{\mathbf{\rho}}}=0
\end{equation}
\noindent where $\bf t$ is the twist of the moving platform and ${\bf \dot{\mathbf{\rho}}}$ is the vector of the active joint velocities.

\begin{figure}[!ht]
\begin{center}
\includegraphics[scale=.50]{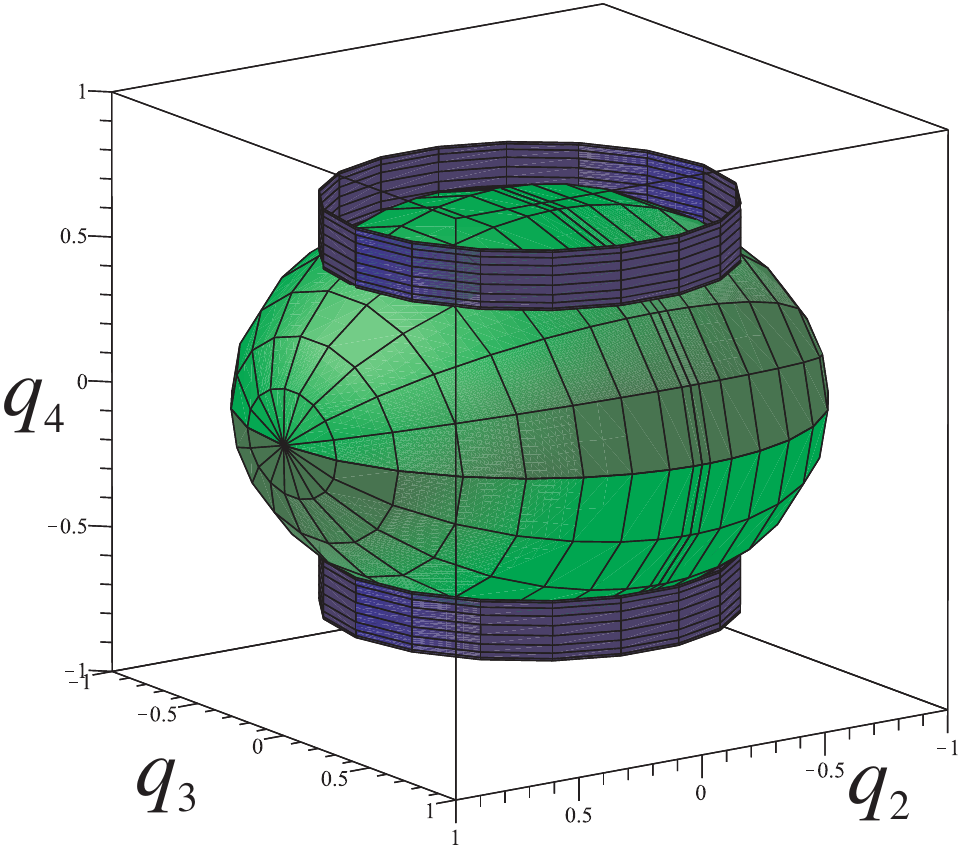}
\end{center}
\caption{Parallel singularity of the 3-\underline{PP}PS robot with quaternion representation}
\label{Fig:Singularity}
\end{figure}

According to the leg topology of the 3-\underline{PP}PS robot, there is no serial singularity because the determinant of the ${\bf B}$ matrix does not vanish. Using the same approach as in \cite{Caro:2010}, we can determine the matrix ${\bf A}$ and its determinant can be factorized as follows:
\begin{equation}
\left( {q_1}^2-{q_2}^2-{q_3}^2+{q_4}^2 \right)  \left( q_1-q_4 \right)  \left( q_1+q_4 \right) =0
\end{equation}
To calculate this result, we use the ``ParallelSingularties'' function from the Siropa library \cite{Siropa}.
To represent this surface, we eliminate $q_1$ thanks to the relation on the quaternions
\begin{equation}
\left( {q_2}^2+{q_3}^2+2 {q_4}^2-1 \right)  \left( 2 {q_2}^2+2 {q_3}^2-1 \right) =0 \label{singularite}
\end{equation}
One of the surface represents a cylinder and the other an ellipsoid. Figure~\ref{Fig:Singularity} depicts these surfaces bounded by the unit sphere.

\begin{figure}[!ht]
\begin{center}
\includegraphics[scale=.50]{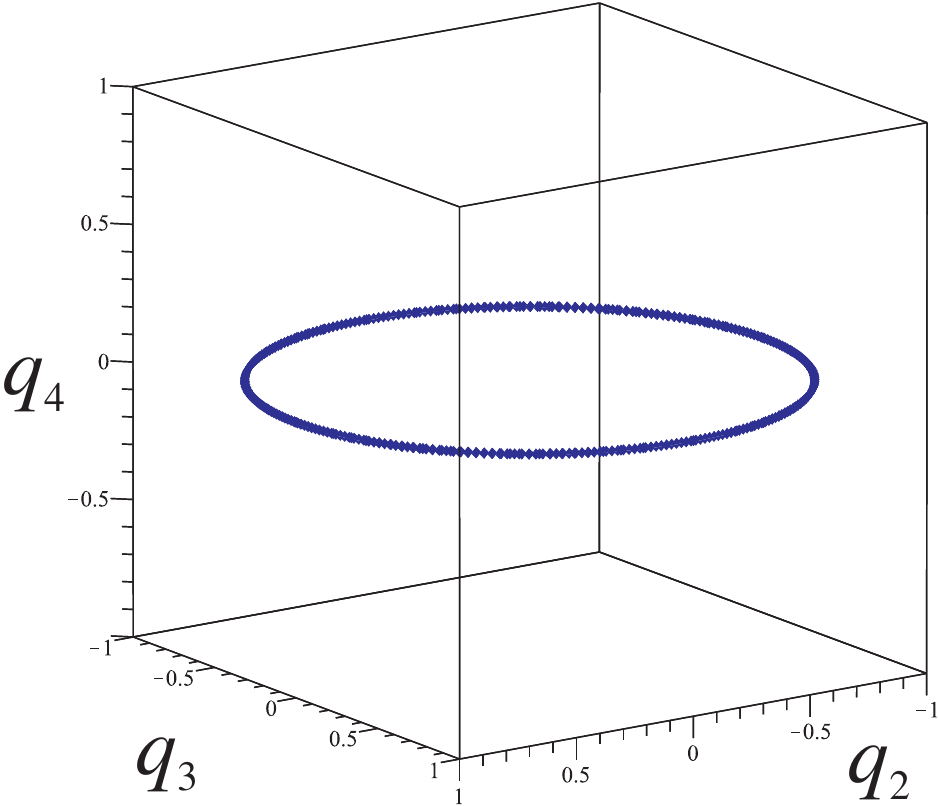}
\end{center}
\caption{Location of self-motions in the workspace.}
\label{Fig:Singularity_self_motion_curve}
\end{figure}

By writing the conditions  $(\rho_{1y}+\rho_{2y}+\rho_{3y} = 0)$ and $\rho_{1z}=\rho_{2z}=\rho_{3z}$ with the constraint equations, the Groebner basis elimination method makes it possible to obtain a set of equations that depends on $q_2$, $q_3$ and $q_4$.
\begin{gather}
q_2 q_4= 0 \nonumber \\
q_3 q_4= 0 \nonumber \\
q_2  \left( {q_2}^2+{q_3}^2-1 \right)=0 \nonumber \\
q_3  \left( {q_2}^2+{q_3}^2-1 \right)=0 \nonumber \\
{q_4}^{3}-q_4 = 0 \nonumber \\
{q_2}^{4}+ \left( {q_3}^2+{q_4}^2-1 \right) {q_2}^2+{q_3}^2{q_4}^2=0
\end{gather}
The intersection of these six equations and the equation~\ref{singularite} is a unit radius circle centered at the origin of the plane $ (0 q_2 q_3)$ as shown in Fig.~\ref{Fig:Singularity_self_motion_curve}. This circle is tangential to the parallel singularity shown in Fig.~\ref{Fig:Singularity}. 

As the singularity does not depend on the position, the self-motion exists for an infinite number of positions of the mobile platform. This phenomenon may not be found if numerical methods are used to solve the DKM because the relations $(\rho_{1y}+\rho_{2y}+\rho_{3y}=0)$ and  $\rho_{1z}=\rho_{2z}=\rho_{3z}$  are never completely satisfied.
\section{Conclusions and Perspectives}
In this article, we studied the parallel robot 3PPPS to explain the presence of a self-motion at the ``home'' position. This movement is a Cardanic movement that has already been studied on the 3-\underline{R}PR parallel robot  or the PamInsa robot. Self-motions can often explain the problems of solving the DKM when using algebraic methods. The calculation of a Groebner base makes it possible to detect this problem but the characterization of this movement for robots with six degrees of freedom is difficult because of the size of the equations. Other robots, such as the CaPaMan at the University of Cassino have the same type of singularity despite the phenomenon never being studied. The objective of future work will be to identify architectures with passive prismatic articulations connected to the mobile platform which is capable of having the same singularity conditions.


\end{document}